\RequirePackage{fix-cm}
\documentclass[twocolumn]{svjour3}          
\smartqed  
\usepackage{graphicx}
\usepackage{amssymb}
\usepackage[cmex10]{amsmath}
\usepackage{array}
\usepackage{url}
\usepackage{color}
\usepackage[tight,footnotesize]{subfigure}
\usepackage{natbib}
\usepackage{hyphenat}
\usepackage[pdftex,
CJKbookmarks=true, bookmarksnumbered=true, bookmarksopen=true,
colorlinks=true,
pdfborder=001,
citecolor=blue, linkcolor=blue, anchorcolor=red, urlcolor=black
]{hyperref}
\usepackage{bbm,dsfont}
\usepackage{indentfirst} 
\usepackage{booktabs}

\usepackage[dvipsnames]{xcolor} 
\usepackage{caption}
\usepackage{subcaption}
\usepackage{multirow}
\usepackage{tabularx}
\usepackage{adjustbox}
\usepackage{pifont}
\usepackage[skip=2pt]{caption}
\usepackage{marvosym}
\usepackage{algorithm}
\usepackage{algorithmic}
\usepackage{amsmath}
\usepackage{rotating}
\usepackage{makecell}

\journalname{}
\definecolor{darkmag}{rgb}{0.55,0,0.55}
\begin{document}
\begin{sloppypar}

\title{Mamba Capsule Routing Towards Part-Whole Relational Camouflaged Object Detection}

\titlerunning{Short form of title}        

\author{Dingwen Zhang \and
        Liangbo Cheng \and
        Yi Liu\textsuperscript{\Letter} \and \\
        Xinggang Wang \and
        Junwei Han\textsuperscript{\Letter}}

\authorrunning{Short form of author list} 

\institute{
Dingwen Zhang \at
Northwestern Polytechnical University, Xi'an 710129, China \\
\email{\href{zhangdingwen2006yyy@gmail.com}{zhangdingwen2006yyy@gmail.com}}
\and
Liangbo Cheng \at
Northwestern Polytechnical University, Xi'an 710129, China \\
\email{\href{lbcheng928@gmail.com}{lbcheng928@gmail.com}}
\and
\textsuperscript{\Letter} Yi Liu\at
Changzhou University, Changzhou 213000, China \\
\email{\href{liuyi0089@gmail.com}{liuyi0089@gmail.com}} 
\and
Xinggang Wang \at
Huazhong University of Science and Technology, Wuhan 430074, China \\
\email{\href{xgwang@hust.edu.cn}{xgwang@hust.edu.cn}}
\and
\textsuperscript{\Letter} Junwei Han \at
Northwestern Polytechnical University, Xi'an 710129, China \\
\email{\href{junweihan2010@gmail.com}{junweihan2010@gmail.com}}
}

\date{Received: date / Accepted: date}

\maketitle

\begin{abstract}
    
The part-whole relational property endowed by Capsule Networks (CapsNets) has been known successful for camouflaged object detection due to its segmentation integrity. However, the previous Expectation Maximization (EM) capsule routing algorithm with heavy computation and large parameters obstructs this trend. The primary attribution behind lies in the pixel-level capsule routing. Alternatively, in this paper, we propose a novel mamba capsule routing at the type level. Specifically, we first extract the implicit latent state in mamba as capsule vectors, which abstract type-level capsules from pixel-level versions. These type-level mamba capsules are fed into the EM routing algorithm to get the high-layer mamba capsules, which greatly reduce the computation and parameters caused by the pixel-level capsule routing for part-whole relationships exploration. On top of that, to retrieve the pixel-level capsule features for further camouflaged prediction, we achieve this on the basis of the low-layer pixel-level capsules with the guidance of the correlations from adjacent-layer type-level mamba capsules. Extensive experiments on three widely used COD benchmark datasets demonstrate that our method significantly outperforms state-of-the-arts. Code has been available on {\href{https://github.com/Liangbo-Cheng/mamba_capsule}{\color{purple}https://github.com/Liangbo-Cheng/mamba\_capsule}}.

\keywords{
Camouflaged object detection \and 
part-whole relationship \and 
capsule network \and mamba}
\end{abstract}

\section{Introduction}

{C}{amouflage} is a form of protective adaptation exhibited by animals in nature, wherein they alter their appearance, texture, and coloration to enhance their ability to hunt for prey and ensure survival. The goal of camouflaged object detection (COD) is to accurately locate and segment concealed objects within their camouflaged surroundings. Thanks to the ability of COD, it has been widely applied in biological conservation \cite{Cai2023,Yang2023}, industrial detection \cite{Rahmon2024,Wang2024}, artistic creation \cite{Zhao2024art,huang2024mrrfs}, and medical image segmentation \cite{Liu2024medical,zhao2021polyp}, \emph{etc}.

Early researchers extract hand-crafted features \cite{Vistnes1989,Fazekas2009,sujitk2013351,Andreopoulos2013}, such as color, texture and optical flow, to detect the camouflaged target. However, these approaches suffer from poor discrimination between foreground and background, resulting in unsatisfactory detection performance. Recently, deep learning based framework has moved forward the development of COD. Especially, since the advent of large-scale COD datasets COD10K \cite{fan20202774}, there have been a large-scale works for COD by simulating the biological mechanisms \cite{fan20202774} and human visual patterns \cite{pang20222160}. They mostly elaborate the indistinguishable feature mining module \cite{chen202226981,sun20211025,mei2023ijcv,luo2023cised} and encompass multiple tasks \cite{he202322046,le201945,sun20221335,zhu20223608} using Convolutional Neural Networks (CNNs) \cite{Bideau2024,zhai20231897,li202110066,Cheng202213864SLT} and Transformers \cite{yang20214126,pei20221937,mei2023ijcv,luo202417169}. However, the strong inherent similarity between the camouflaged object and its background restricts the feature extraction capability of both CNN and Transformer networks that try to find discriminative regions, causing incomplete detection easily with object details missed or local parts lost.

To cater to this issue, part-whole relational property endowed by Capsule Networks (CapsNets) \cite{liu20215154} has been proven successful for the complete segmentation of camouflaged object, which is implemented by excavating the relevant parts of the object \cite{liu20191232}. However, the  previous Expectation-Maximization (EM) routing \cite{hinton2018matrix} makes the part-whole relational COD \cite{liu20215154} challenging in terms of computational complexity, parameter, and inference speed. The reason behind is that the previous pixel-level EM routing inevitably generates large-scale capsule assignments at the pixel level, causing large-scale dense computation.

Recently, Vision Mamba (VMamba) \cite{liu2024arxiv} has successfully adapted the mamba \cite{gu20242312} that is renowned for efficient modeling of long sequences to address computer vision tasks. Specifically, four-direction scans are implemented on the 2D input to obtain 1D sequence tokens for further image recognition, which are fed into the selective State Space Models (SSMs) \cite{gu2022efficiently,GuJasion2021} to attend the important tokens. During the selective SSMs stage, the 1D sequence latent state implicitly models the global context. Besides, four-direction scans in VMamaba ensure spatial context for the latent state. 

In this paper, inspired by VMamba \cite{liu2024arxiv}, we introduce VMamba in the task of part-whole relational COD with the aim of designing a lightweight capsule routing. To this end, we propose a novel Mamba Capsule Routing Network (MCRNet) to detect the camouflaged object. Specifically, the 2D pixel-level spatial capsules are first fed into the scanning mechanism to obtain 1D capsule tokens in various scanning direction, which are input into the selective SSM \cite{gu20242312} to learn the 1D implicit latent state capsules, named mamba capsules. Such Mamba Capsule Generation (MCG) module ensures to extract the 1D type-level capsules from the pixel-level 2D primary capsules. The EM routing \cite{hinton2018matrix} absorbs the mamba capsules to generate the high-layer\footnote{High-layer and whole-level can be applied in an interleaved manner.} versions and the type level, which implements the part-whole relationships with lightweight routing computation. On top of that, to retrieve the pixel-level spatial details from the high-layer mamba capsules for the final dense prediction of camouflaged object, we design a Capsules Spatial Details Retrieval (CSDR) module. Concretely, we rely on two components, including the low-layer\footnote{Low-layer and part-level can be applied in an interleaved manner.} pixel-level capsules and the correlation of adjacent-layer mamba capsules, to achieve the high-layer capsule spatial details. Using this mechanism, spatial capsules in four scanning directions are integrated to be in the uniform direction for the final camouflage detection.

\begin{figure}
\centering
\includegraphics[width=0.48\textwidth]{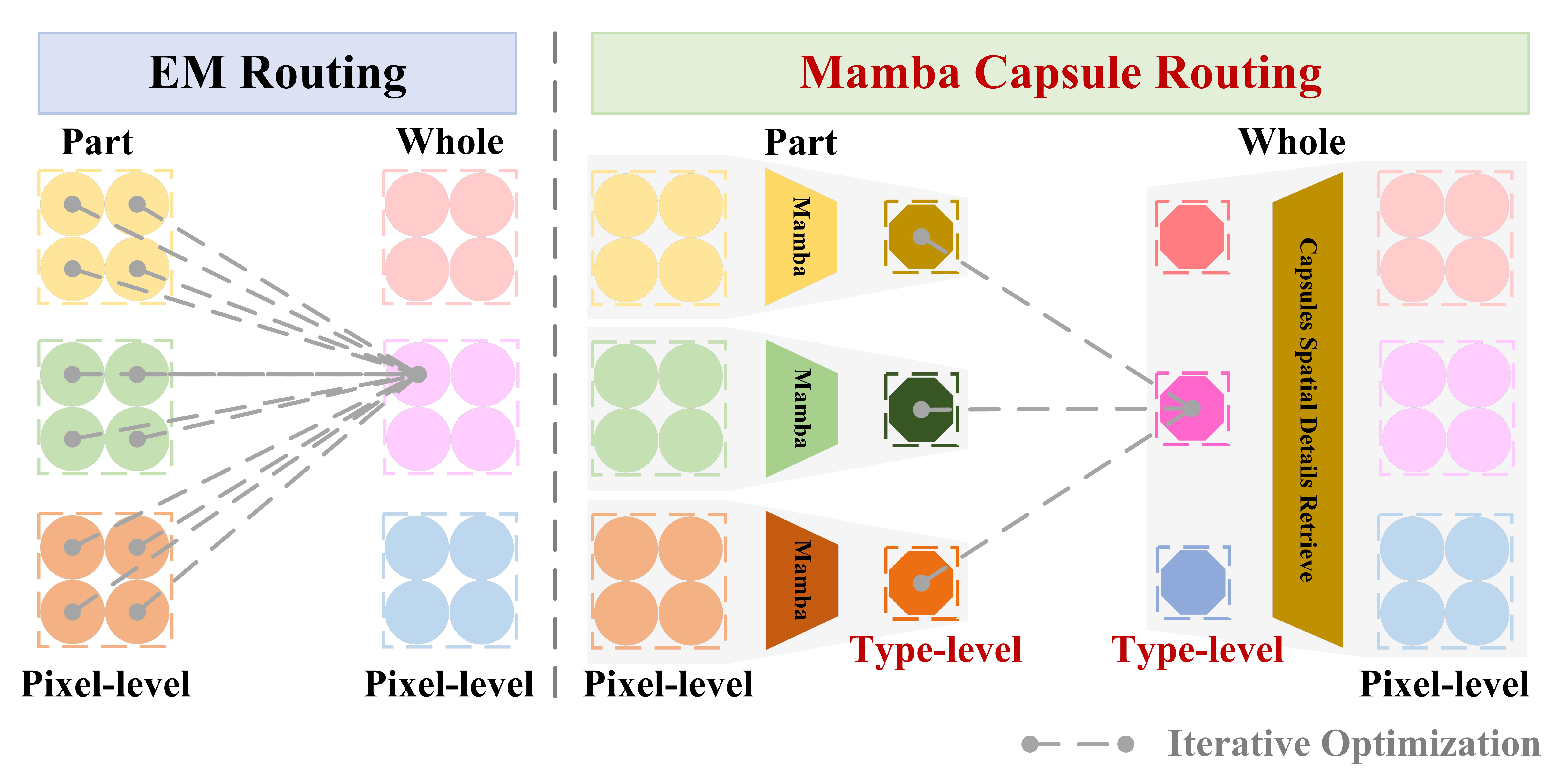}
\caption{
Different capsule routings for part-whole relational camouflaged object detection. The original EM routing \cite{hinton2018matrix} involves a significant number of parameters and routing complexity at the pixel level due to the dense routing. Differently, our proposed MCRNet compresses spatially pixel-level capsules into type-level capsules, leading to a substantial complexity-reduction type-level capsule routing. On top of that, the capsules spatial details retrieval is used to learn the spatial details of mamba capsules for further camouflaged object detection.}
\label{motivation}
\vspace{-3mm}
\end{figure}

To sum up, the contributions of this paper are as follows:

\begin{itemize} 

\item We design a novel MCRNet for the part-whole relational COD task, which gets the capsule routing complexity reduced significantly. To the best of our knowledge, it is the first attempt to employ mamba for CapsNets and the the task of COD.

\item We purpose a MCG  module to generate the type-level mamba capsule from the pixel-level capsules, which helps for the lightweight capsule routing.

\item We design a CSDR module to retrieve the spatial details from the high-layer type-level mamba capsules for the dense detection of camouflaged object.

\item Comprehensive experiments demonstrate that our proposed MCRNet achieves superior performance on three widely-used COD datasets compared to 25 existing state-of-the-art methods.

\end{itemize}

This paper is organized as follows. Sec. \ref{sec:Related} reviews the related references to our work. Sec. \ref{sec:Preliminaries} summarizes the preliminaries of mamba. Sec. \ref{sec:Proposed} describes the details of the proposed MCRNet. Sec. \ref{sec:Experiment} carries out abundant experiments and analyses to understand our method. Sec. \ref{sec:Conclusion} concludes the paper.

\section{Related Work}
\label{sec:Related}

In this section, we will review references related to our work, including camouflaged object detection, capsule network, and vision mamba.

\subsection{Camouflaged Object Detection}

The task of COD refers to accurately segmenting objects that are intentionally designed or naturally evolved to blend into their surroundings, which is rather challenging due to the high similarity between target object and background. Researchers have done a lot of wonderful works that have greatly advanced the development. In the early stage, most of the  works are developed based on the hand-crafted features, \emph{e.g.}, colour \cite{Huerta2007improving}, 3D convexity \cite{Pan20113dconvexity} and intensity features \cite{sengottuvelan2008intensity}. However, they are relatively less robust and prone to fail in complex scenarios with low contrast situations. With the popularity of deep learning \cite{fan20202774}, recent works focus on mining more detailed features in a learning manner to distinguish camouflaged objects from their surroundings. Inspired by the biological mechanisms in nature or human visual psychological patterns, Fan \emph{et al.} \cite{fan20226024} mimiced the behavior process of predators to simulate the search and identification towards preys. Pang \emph{et al.} \cite{pang20222160} adopted the zoom mechanism of humans when they observed fuzzy objects for the task of COD. Jia \emph{et al.} \cite{jia20224703} segmented, magnified and reiterated the camouflaged object in a coarse-to-fine manner with the multi-stage strategy. Besides, some feature mining modules are elaborated for unearthing the subtle discriminative features of camouflaged objects. For example, Zhu \emph{et al.} \cite{zhu20213599} focused on  texture-aware learning. Mei \emph{et al.} \cite{mei20218772} aimed to learn contextual-aware information. He \emph{et al.} \cite{he202322046} and Zhong \emph{et al.} \cite{zhong20224494} introduced frequency clues to aid camouflaged object detection. Moreover, incorporating auxiliary tasks with the COD task can facilitate the precise segmentation map, such as classification \cite{le201945}, edge/boundary detection \cite{sun20221335,zhu20223608,luo202417169} and object ranking \cite{lv202111586}. The methods above, which are based on CNNs and Transformer, exhibit suboptimal performance in detecting camouflaging objects with high similarity and low contrast to their background. Alternatively, Liu \emph{et al.} \cite{liu20215154} made the first attempt to complete COD task in the part-whole relational perspective successfully. 

In this paper, in along with the pipeline of the part-whole relational COD, our work takes a further step in terms of the lightweight capsule routing, which is achieved by introducing VMamba \cite{liu2024arxiv} to CapsNets \cite{hinton2018matrix}.

\subsection{Capsule Network}

The history of part-whole representation goes back several decades. Krivic and Solina \emph{et al.} \cite{krivic20041077} recognized articulated objects based on part-level descriptions obtained by the Segmentor system \cite{chen1993129}. Girshick \emph{et al.} \cite{girshick2015437} designed a CNN to formulate the deformable part model using a distance transform pooling, object geometry filters, and maxout units. To address the problem of CNNs with space invariance, Hinton \emph{et al.} \cite{hinton201144,sabour20173859,hinton2018matrix} explored the part-whole relationships by the CapsNets, which route low-level capsules (parts) to their familiar high-level ones (wholes). As the previous EM routing \cite{hinton2018matrix} consumes too much computational resources, lots of improved works focus on the lightweight routing. Liu \emph{et al.} \cite{liu20226719} disentangled two orthogonal 1D routings, which greatly reduce parameters and routing complexity, resulting in faster inference than the previous omnidirectional 2D routing adopted by the EM routing strategy. Liu \emph{et al.} \cite{liu20241} presented a residual pose routing. Likewise, Geng \emph{et al.} \cite{geng20246037} designed an orthogonal sparse attention routing to reduce redundancy and reducing parameters. 

Despite the above improvement for capsule networks has made great progress in reducing the redundancy, the EM routing \cite{hinton2018matrix} still remains the pixel level, resulting in large-scale capsule assignments and computational complexity. Unlike previous works, we introduce the VMamba \cite{liu2024arxiv} to generate type-level mamba capsules from the pixel-level capsules for routing, which ensures a lightweight computation.

\begin{figure*}
\centering
\includegraphics[width=1.0\textwidth]{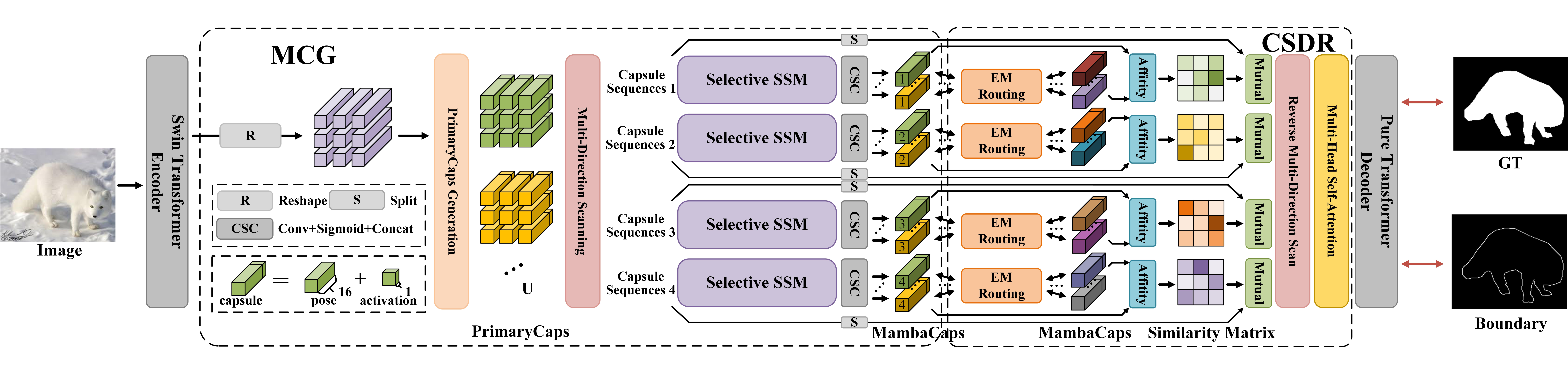}\\
\caption{The overall architecture of MCRNet. The long-range context from Swin Transformer \cite{liu20219992} is first fed into the designed MCG module. In MCG, each type of constructed primary capsules is scanned in four directions, which are further input into the selective SSM \cite{gu20242312} module to achieve the implicit latent state, which is treated as the type-level mamba capsules for subsequent routing to learn the high-layer mamba capsules. In the following, the proposed CSDR module is used to retrieve the spatial details of mamba capsules for final camouflaged prediction. To learn primitive object edges, the object boundary label is also taken into account for training.
}\label{fig:sub}
\vspace{-3mm}
\end{figure*}

\subsection{Vision Mamba}

Considering that the high-order complexity of the self-attention mechanism in the transformer increases quadratically with increasing image size, mamba \cite{gu20242312} has recently shown good performance in long sequence modeling and can be a promising alternative. Due to the low complexity, mamba has been involved in the computer vision community. For example, Zhu \emph{et al.} \cite{zhu2024arxiv} designed the first mamba-based backbone network to generate a linear computational complexity while retaining advantages of vision transformer, which showcases superior performance and the ability to capture complex visual dynamics. Liu \emph{et al.} \cite{liu2024arxiv} designed a cross-scan mechanism to bridge the gap between 1D array
scanning and 2D plain traversing. Ma \emph{et al.} \cite{ma2024umamba} proposed a hybrid CNN-SSM structure to capture local fine-grained features and remote dependencies in images to solve the problem of biomedical image segmentation. Liang \emph{et al.} \cite{liang2024pointmam} introduced a reordering strategy to scan data in a specific sequence to capture point cloud structures. 

In this paper, inspired by the computational efficiency, we introduce VMamba \cite{liu2024arxiv} in CapsNets \cite{hinton2018matrix} to solve the part-whole relational COD task, which generates the type-level mamba capsules from the pixel-level versions using the implicit hidden state for further capsules routing.

\begin{figure*}
\centering
\includegraphics[width=1.0\textwidth]{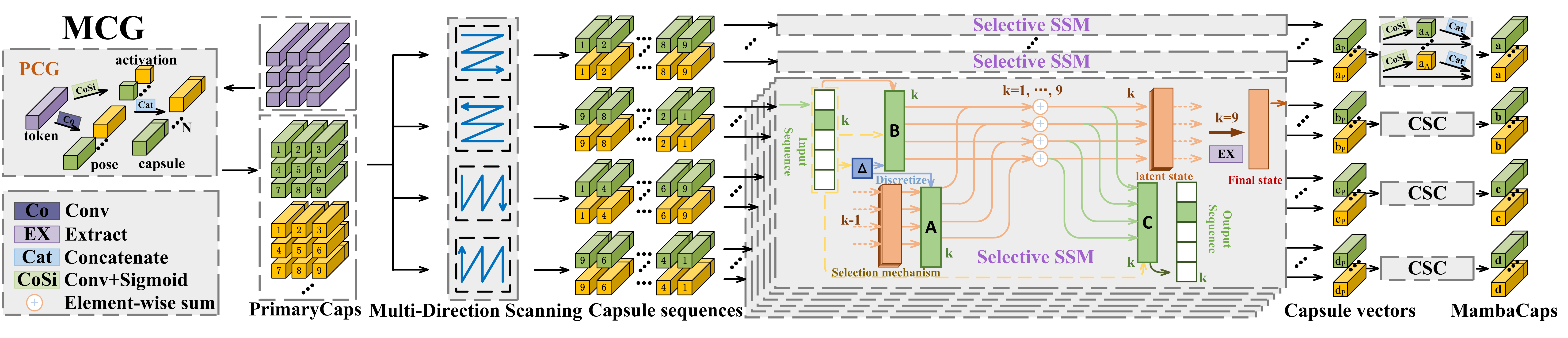}\\
\caption{Details of MCG. The generated primary capsules are scanned in different directions into capsule sequences, which are input to selective SSM \cite{gu20242312} module. The final latent state is chosen as mamba capsules vectors.
}\label{fig:MCG}
\vspace{-3mm}
\end{figure*}

\section{Preliminaries}
\label{sec:Preliminaries}

In this section, we will review the mechanism of SSMs \cite{gu20242312,liu2024arxiv} with details, which will be involved in the proposed MCG module.

The original SSMs \cite{kalman1960} are regarded as Linear Time Invariant (LTI) systems that map the input stimulation $\mathbf{x}(t) \in \mathbb{R}$ to response $\mathbf{y}(t) \in \mathbb{R}$ through the latent state $\mathbf{h}(t) \in \mathbb{R}^{N}$, which is expressed in linear ordinary differential equations
\begin{equation}
\label{equ:ssm}
\begin{array}{l}
\mathbf{h}^{'}(t) = \mathbf{A}\mathbf{h}(t) + \mathbf{B}\mathbf{x}(t), \\ \ \mathbf{y}(t) = \mathbf{C}\mathbf{h}(t),
\end{array}
\end{equation}where $\mathbf{A} \in \mathbb{R}^{N \times N}$ means the evolution parameter. $\mathbf{B} \in \mathbb{R}^{N \times 1}$ and $\mathbf{C} \in \mathbb{R}^{1 \times N}$ are the projection parameters. 

To facilitate integration into the deep learning model, the continuous system is discretized, including a time-scale parameter $\mathbf{\Delta}$ to transform the continuous parameters $\mathbf{A}$, $\mathbf{B}$ to discrete parameters $\overline{\mathbf{A}}$, $\overline{\mathbf{B}}$. Using the zero-order hold (ZOH) relu:
\begin{equation}
\label{equ:discretization}
\begin{array}{l}
\overline{\mathbf{A}} = \mathrm{exp}(\mathbf{\Delta}\mathbf{A}),\\
\overline{\mathbf{B}} = (\mathbf{\Delta}\mathbf{A})^{-1}(\mathbf{\Delta}\mathbf{A}-\mathbf{I}) \cdot \mathbf{\Delta}\mathbf{B}.
\end{array}
\end{equation}
After the discretization, Eq. $(\ref{equ:ssm})$ can be rewritten as
\begin{equation}
\label{equ:s4rnn}
\begin{array}{l}
\mathbf{h}_{t} = \overline{\mathbf{A}}\mathbf{h}_{t-1} + \overline{\mathbf{B}}\mathbf{x}_{t}, \\  \mathbf{y}_{t} = \mathbf{C}\mathbf{h}_{t}.
\end{array}
\end{equation}
Finally, the model calculates the output by global convolution
\begin{equation}
\label{equ:s4cnn}
\begin{array}{l}
\overline{\mathbf{K}} = (\mathbf{C}\overline{\mathbf{B}}, \mathbf{C}\overline{\mathbf{A}}\overline{\mathbf{B}}, \dots, \mathbf{C}{\overline{\mathbf{A}}}^{L-1}\overline{\mathbf{B}}), \\  \mathbf{y} = \mathbf{x} \ast \overline{\mathbf{K}},
\end{array}
\end{equation}where $L$ is the length of the input sequence $\mathbf{x}$, and  $\overline{\mathbf{K}} \in \mathbb{R}^{L}$ is a structured convolutional kernel. $\ast$ indicates the operation of convolution.

To tackle the limitation of LTI SSMs (Eq. $(\ref{equ:ssm})$) in capturing the contextual information, Gu \emph{et al.} \cite{gu20242312} proposed a novel parameterization method for SSMs that integrates an input-dependent selection mechanism, \emph{i.e.}, 
\begin{equation}
\label{equ:selective}
\begin{array}{l}
\mathbf{B}=\mathrm{{Linear}}_{N}(\mathbf{x}),\, \mathbf{C}=\mathrm{{Linear}}_{N}(\mathbf{x}),\, \mathbf{\Delta}= \mathrm{{Linear}}_{D}(\mathbf{x}), 
\end{array}
\end{equation}where $\mathrm{{Linear}}_{N}$ and $\mathrm{{Linear}}_{D}$ are the parameterized projection to dimension $N$ and $D$, respectively. Eq. $(\ref{equ:selective})$ focuses on or ignores specific tokens selectively according to the input sequence, making the model process information efficiently.

\section{Proposed Method}
\label{sec:Proposed}

In this section, we will illustrate the details of MCRNet for camouflaged object detection.

\subsection{Overview}

Fig. \ref{fig:sub} depicts the overall architecture of our proposed MCRNet, including a Swin Transformer encoder \cite{liu20219992}, a Mamba Capsule Generation (MCG) module, a Capsules Spatial Details Retrieval (CSDR) module, and a multi-task learning decoder \cite{liu20214702}. To be specific, the input image is divided into non-overlapped patches after data augmentation, which are fed into Swin Transformer \cite{liu20219992} to capture long-range context features. On top of that, MCG is designed to generate type-level mamba capsules from the pixel-level capsules for further capsule routing to obtain the high-level mamba capsules. To learn the spatial details of mamba capsules, CSDR is developed to retrieve the spatial resolution of the high-layer type-level mamba capsules. Finally, a multi-task learning decoder including camouflaged detection and edge detection is designed to detect the camouflaged object with excellent boundary.

\subsection{Transformer Encoder}

The transformer encoder first splits the input RGB image \(\mathbf{I} \in \mathbb{R}^{C \times H \times W}\) into non-overlapped patches (\(p \times p\)) by a patch embedding module, where \(C\), \(H\) and \(W\) denote channel size, height and width of image \(\mathbf{I}\), respectively, and \(p\) = 16. These image patches are linearly projected into a 1D sequence of token embeddings \(\mathbf{F}^{E} \in \mathbb{R}^{l \times d}\), where \(l = HW/p^{2}\) and  \(d\) are the length of the patch sequence and the channel dimension, respectively. The Swin Transformer \cite{liu20219992} encoder is used to capture global dependencies \(\mathbf{F}_{i}^{E} \in \mathbb{R}^{l_{i} \times d_{i}}\), where \(i \in \lbrack 0,1,2,3\rbrack\) indicates the index of blocks in the encoder, $l_{i}$ and $d_{i}$ mean the length of the sequence and the channel dimension of the token, respectively. Its unique shifted windowing mechanism reduces the computational burden with more efficient batch computation, showing efficiency and high performance.

\subsection{Mamba Capsule Generation}
\label{sec:Network}

In this subsection, we will detail the MCG module that learns the type-level mamba capsules from the pixel-level capsules, which is composed by primary capsules generation, multi-direction serialization, implicit latent state learning and mamba capsule acquisition.

\textbf{Step 1:} Primary capsules generation. As shown in Fig. \ref{fig:MCG}, the feature sequence \(\mathbf{F}_{2}^{E} \in \mathbb{R}^{l_{2} \times d_{2}}\) obtained by the encoder is reshaped into  \(\mathbf{F}^{'} \in \mathbb{R}^{h_{2} \times w_{2} \times d_{2}}\) to facilitate subsequent Primary Capsules (PrimaryCaps) generation \(\mathbf{P} \in \mathbb{R}^{h_{2} \times w_{2} \times {O} \times U}\), which contains the pose matrix \(\mathbf{P}_{pose} \in \mathbb{R}^{h_{2} \times w_{2} \times {O}_{P} \times U}\) and the activation value  \(\mathbf{P}_{act} \in \mathbb{R}^{h_{2} \times w_{2} \times {O}_{A} \times U}\), where $O=\{{O}_{P}=16, {O}_{A}=1\}$ is the dimension of the pose matrix and the activation, \(U\) represents the number of primary capsules, \emph{i.e.}, 
\begin{equation}
\label{equ:pcg}
\begin{split}
\mathbf{P} = \mathrm{Cat}(\mathbf{P}_{pose},  \mathbf{P}_{act}) =\mathrm{Cat}(\Phi (\mathbf{F}^{'}), \mathrm{Sigmoid}(\Phi (\mathbf{F}^{'}))),
\end{split}
\end{equation}where $\Phi(\cdot)$ represents the operation of convolution, batch normalization and relu. $\mathrm{Sigmoid}(\cdot)$ means the sigmoid function. $\mathrm{Cat}(\cdot)$ represents the concatenation.

\textbf{Step 2:} Multi-direction serialization. Using the scanning mechanism of VMamba \cite{liu2024arxiv} for providing more accurate and rich 2D spatial context, the 2D primary capsules are serialized to four groups of 1D capsule sequences $\mathbf{S}_{g} = \{ \mathbf{S}_{1}, \dots, \mathbf{S}_{G} \} \in \mathbb{R}^{V \times O \times U }$, where $G=4$ means four various scanning directions as shown in Fig. \ref{fig:MCG} (positive Z shape, inverted Z shape, positive N shape, and inverted N shape), $U$ represents the number of capsule sequences, $V = h_{2} \times w_{2}$ is the length of the capsule sequence and $O$ means the dimension of the capsule token. In a certain scanning direction $g$, each capsule sequence $ \mathbf{S}_{g}(u) = \{ \mathbf{S}_{g}(1), \dots, \mathbf{S}_{g}(U) \} \in \mathbb{R}^{V \times O}$ represents various part object. In a certain capsule sequence $\mathbf{S}_{g}(u)$, there are $V$ capsule tokens $ \mathbf{S}_{g}(u,v) = \{ \mathbf{S}_{g}(u,1), \dots, \mathbf{S}_{g}(u,V) \} \in \mathbb{R}^{O}$.

\textbf{Step 3:} Implicit latent state learning. During the selective SSM, the current latest implicit latent state is associated with both the accumulated latent state and the current input token, which can be formulated as
\begin{equation}
\setlength\abovedisplayskip{6pt}
\setlength\belowdisplayskip{6pt}
\label{equ:ssm1}
\begin{array}{l}
\mathbf{h}_{g}(u,v) = \overline{\mathbf{A}} \mathbf{h}_{g}(u,v-1) + \overline{\mathbf{B}} \mathbf{S}_{g}(u,v),
\end{array}
\end{equation}
where $\mathbf{S}_{g}(u,v)$ represents the ${v}^{th}$ token in the ${u}^{th}$ capsule sequence obtained in the scanning direction  $g$. $\mathbf{h}_{g}(u,v) \in \mathbb{R}^{N}$ means the updated implicit latent state after the token $\mathbf{S}_{g}(u,v)$ is input. \(\overline{\mathbf{A}}\) and \(\overline{\mathbf{B}}\) represent the discretized evolution parameters of the model, which will be computed based on the input sequence. 

The most recent output mamba token can be obtained by utilizing this latest latent state
\begin{equation}
\setlength\abovedisplayskip{6pt}
\setlength\belowdisplayskip{6pt}
\label{equ:ssm2}
\begin{array}{l}
\mathbf{F}_{g}^{M}(u,v) = \mathbf{C} \mathbf{h}_{g}(u,v),
\end{array}
\end{equation}
where $\mathbf{F}_{g}^{M}(u,v)$ indicates the output mamba token corresponding to $\mathbf{S}_{g}(u,v)$. \(\mathbf{C}\) represents the projection parameter of the model, which is computed relying on the input sequence. 

As each token $\mathbf{S}_{g}(u,v)$ in the sequence $\mathbf{S}_{g}(u)$ is fed into the selective SSM, the implicit latent state $\mathbf{h}_{g}(u,v)$ is updated constantly. The final implicit latent state $\mathbf{h}_{g}(u,V)$, which is defined as mamba capsule vectors completes the global modeling of the sequence $\mathbf{S}_{g}(u)$. Algorithm \ref{alg:Selective SSM} lists the process of mamba capsule vectors learning in details.

\begin{algorithm}[t]
\caption{\textbf{Mamba Capsule Vectors Learning.} \(\mathbf{S}\)
is the input capsule sequence. \(\mathbf{F}^{M}\) is the output mamba sequence. \(\mathbf{h}\) is the mamba capsule vectors, which is also the implicit latent state from the last iteration.}
\label{alg:Selective SSM}
\begin{algorithmic}
\STATE {\textbf{Procedure:}\\
\ \ \ \ 1. Initialize parameter \textbf{A}: \emph{(D, N)}\\
\ \ \ \ 2. Learn parameters \textbf{B}, \textbf{C}, \(\mathbf{\Delta}\):\\
\ \ \ \ \big| \ \ \ \ \textbf{B}: \emph{(B, L, N)} ← \(\mathrm{{Linear}}_{N}(\mathbf{S})\)\\
\ \ \ \ \big| \ \ \ \ \textbf{C}: \emph{(B, L, N)} ← \(\mathrm{{Linear}}_{N}(\mathbf{S})\)\\
\ \ \ \ \big| \ \ \textcolor{gray}{/*\(\mathrm{Linear}_{N}\) is a linear projection to dimension \(N\)*/}\\
\ \ \ \ \big| \ \ \ \ \(\mathbf{\Delta}\): \emph{(B, L, D)} ← \(\mathrm{softplus}\)( $\Omega$\\
\ \ \ \ \big| \ \ \ \ \ \ \ \ \ \ \ \ \ \ \ \ \ \ \ \ \ \ \ \ + \(\mathrm{Broadcast_{D}({Linear}_{1}}(\mathbf{S}))\)\\
\ \ \ \ \big| \ \ \textcolor{gray}{/*softplus ensures activation*/}\\
\ \ \ \ \big| \ \ \textcolor{gray}{/* $\Omega$ means initialize parameters*/}\\
\ \ \ \ 3. Discretization parameter \(\overline{\mathbf{A}}\)\textbf{,}
  \(\overline{\mathbf{B}}\):\\
\ \ \ \ \big| \ \ \ \ \(\overline{\mathbf{A}},\overline{\mathbf{B}}\mathbf{:}\) \emph{(B, L, D,N)} ← discretize(\(\mathbf{\Delta}\), \textbf{A}, \textbf{B})\\
\ \ \ \ \big| \ \ \textcolor{gray}{/*Eq. (\ref{equ:discretization})*/}\\
\ \ \ \ 4. Get mamba capsule vectors $\mathbf{h}$ and sequence $\mathbf{F}_{M}$:\\
\ \ \ \ \big| \ \ \ \ \(\mathbf{h}\): (B, D, N) ← SSM(\(\overline{\mathbf{A}}\)\textbf{,}
\(\overline{\mathbf{B}}\)\textbf{, C})(\(\mathbf{S}\))\\
\ \ \ \ \big| \ \ \ \ \(\mathbf{F}^{M}\): (B, L, D) ← SSM(\(\overline{\mathbf{A}}\)\textbf{,}
\(\overline{\mathbf{B}}\)\textbf{, C})(\(\mathbf{S}\))\\
\ \ \ \ \big| \ \ \textcolor{gray}{/*Eq. (\ref{equ:ssm1}) and Eq. (\ref{equ:ssm2})*/}\\
\ \ \ \ 5. Return \(\mathbf{h}\), \(\mathbf{F}^{M}\).\\}
\end{algorithmic}
\end{algorithm}

\textbf{Step 4:} Mamba capsule acquisition. Ultimately, in the certain scanning direction $g$, we obtain learned mamba sequences $\mathbf{F}_{g}^{M} \in \mathbb{R}^{V \times O \times U}$,  and the implicit latent state $\mathbf{h}_{g} \in \mathbb{R}^ {N \times U}$. Due to the fact that the final latent state $\mathbf{h}_{g}(u,V)$ implicitly explores the global context of the corresponding sequence, we choose it as the capsule pose vector $\mathbf{M}_{pose,g} \in \mathbb{R}^ {N \times U}$, which has a comprehensive understanding about the pixel-level capsules. Based on the pose vector $\mathbf{M}_{pose,g}$, we can compute the activation values $\mathbf{M}_{act,g} \in \mathbb{R}^{O_{A} \times U}$ through
\begin{equation}
\label{equ:mambaact}
\begin{array}{l}
\mathbf{M}_{act,g} = \mathrm{Sigmoid}(\Phi (\mathbf{M}_{pose,g})).
\end{array}
\end{equation} To this end, the type-level Mamba Capsules (MambaCaps) \(\mathbf{M}_{g} \in \mathbb{R}^{1 \times 1 \times O \times U}\) is constructed as 
\begin{equation}
\label{equ:mambacaps}
\begin{array}{l}
\mathbf{M}_{g} = \mathrm{Unsqueeze}(\mathrm{Cat}(\mathbf{M}_{pose,g}, \mathbf{M}_{act,g})),
\end{array}
\end{equation}where $\mathrm{Unsqueeze}(\cdot)$ represents the operation of unsqueeze. In Eq. (\ref{equ:mambacaps}), the type-level mamba capsules are derived from the pixel-level capsules while preserving global context, which helps to get routing computation reduced significantly.

\begin{figure}[t]
\centering
\includegraphics[width=1\linewidth]{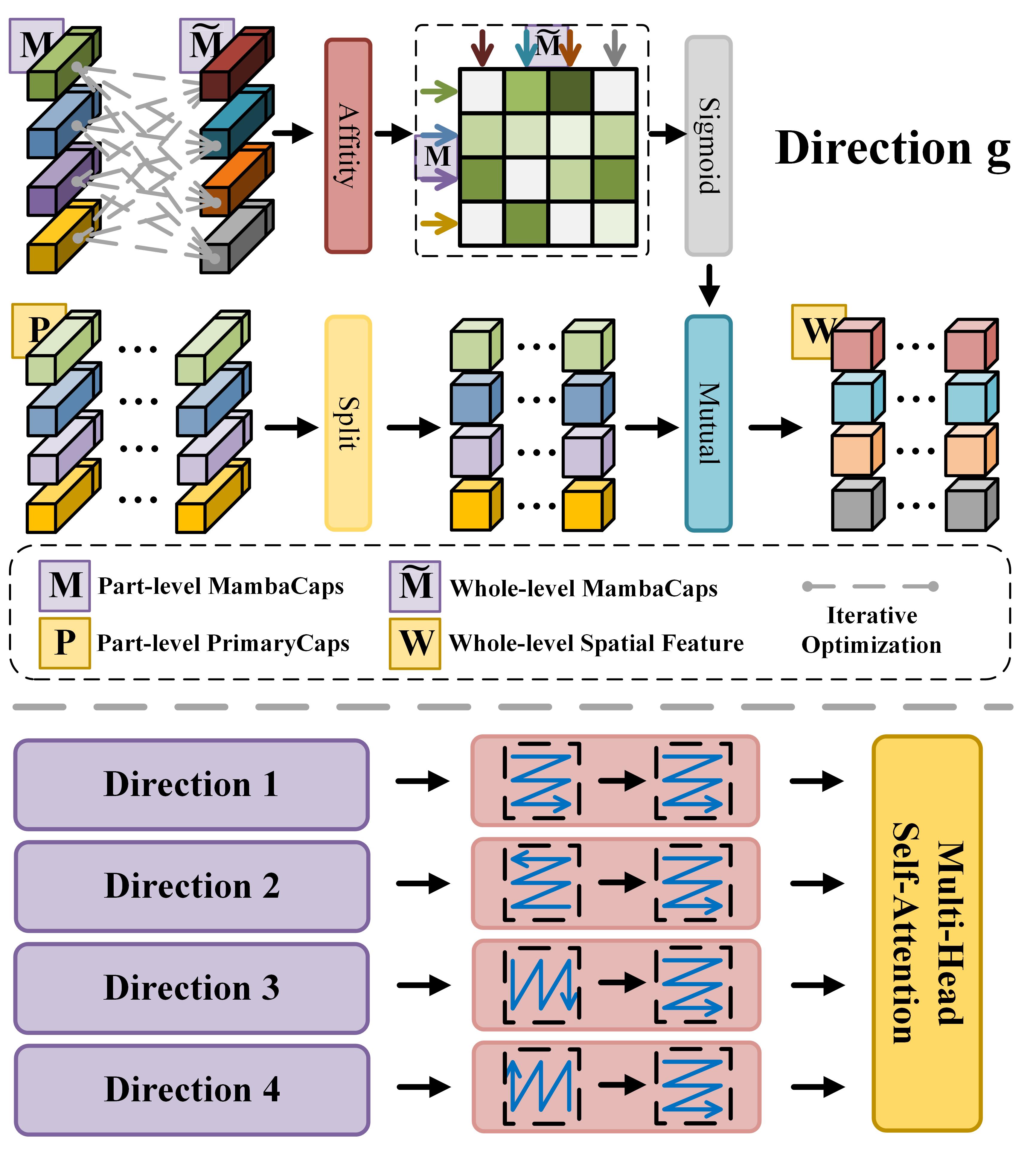}
\caption{Details of CSDR. Adjacent-layer mamba capsules compute their correlation, which is integrated with the low-layer pixel-level capsules to achieve the spatial details of the high-layer mamba capsules. The bottom indicates that four scanning directions will be transformed into a uniform direction to fuse the spatial details of the high-layer mamba capsules in different scanning directions through multi-head self-attention.}
\label{fig:SCR}
\end{figure}

\subsection{Capsules Spatial Details Retrieval}

In this subsection, we will detail the CSDR module for further camouflaged prediction, which is composed by high-layer mamba capsules learning, adjacent-layer mamba capsules correlation 
and high-layer capsules spatial details retrieval.

\textbf{Step 1:} High-layer mamba capsules learning. As shown in Fig. \ref{fig:SCR}, under the certain scanning direction $g$, the obtained mamba capsules are fed into EM routing \cite{hinton2018matrix} with iterative refinement for mining the part-whole relationship at the type-level, and the high-layer mamba capsules can be computed via
\begin{equation}
\label{equ:mambaem}
\begin{array}{l}
\lbrack \widetilde{\mathbf{M}}_{g}(1),\ldots,\widetilde{\mathbf{M}}_{g}(U) \rbrack = \mathrm{EM}\lbrack \mathbf{M}_{g}(1),\ldots,\mathbf{M}_{g}(U)\rbrack,
\end{array}
\end{equation}where $\mathrm{EM}(\cdot)$ represents the EM routing algorithm \cite{hinton2018matrix} and \(\widetilde{\mathbf{M}}_{g}= \lbrack \widetilde{\mathbf{M}}_{g}(1),\ldots,\widetilde{\mathbf{M}}_{g}(U)\rbrack \in \mathbb{R}^{1 \times 1 \times O \times U}\) means $U$ explored high-layer mamba capsules under the certain scan direction $g$.

\textbf{Step 2:} Adjacent-layer mamba capsules correlation. To utilize the type-level mamba capsules for the dense prediction of camouflaged object, it is necessary to retrieve the spatial details of the mamba capsules. 

Under the certain scanning direction $g$, the cosine similarity matrix \(\mathbf{E}_{g}(m,n) \in \mathbb{R}^{U \times U}\) depicts the similarity degree between the adjacent-layer type-level mamba capsules \(\mathbf{M}_{g}(m)\) and \(\widetilde{\mathbf{M}}_{g}(n)\), where $m \in \lbrack 1, \dots, {U}\rbrack$ is the row index of the matrix $\mathbf{E}_{g}$ and the index of $\mathbf{M}_{g}$, $n \in \lbrack 1, \dots, {U}\rbrack$ represents the column index of the matrix $\mathbf{E}_{g}$ and the index of $\widetilde{\mathbf{M}}_{g}$, which can be computed by
\begin{equation}
\label{equ:similaritymatrix}
\begin{split}
\mathbf{E}_{g}(m,n) = \frac{\sum_{x = y = 1}^{O}{\mathbf{M}_{g}(m,x) \times \widetilde{\mathbf{M}}_{g}(n,y)}}{\sqrt{\sum_{x = 1}^{O}{(\mathbf{M}_{g}(m,x))^{2}}}\sqrt{\sum_{y = 1}^{O}{(\widetilde{\mathbf{M}}_{g}(n,y))^{2}}}},
\end{split}
\end{equation}where $x \in \lbrack 1, \dots, O\rbrack$, $y \in \lbrack 1, \dots, O\rbrack$ represent the element index of \(\mathbf{M}_{g}(m)\) and \(\widetilde{\mathbf{M}}_{g}(n)\), respectively.

To further enhance the differentiation and distinguishable ability of the cosine similarity between adjacent-layer mamba capsules, the sigmoid function is utilized for activation as
\begin{equation}
\label{equ:activation}
\begin{array}{l}
\mathbf{\widehat{E}}_{g}(m,n) = \mathrm{Sigmoid}(\mathbf{E}_{g}(m,n)).
\end{array}
\end{equation}

In the cosine similarity matrix $\mathbf{\widehat{E}}_{g}$, the element in row \(m\) and column \(n\) represents the degree of correlation between the ${m}^{th}$ low-layer mamba capsule and the ${n}^{th}$ high-layer mamba capsule, which also reflects the correlation between the corresponding pixel level adjacent layer capsules.

\textbf{Step 3:} High-layer capsules spatial details retrieval. Under the certain scanning direction $g$, the activation values \(\mathbf{S}_{act,g} \in \mathbb{R}^{V \times O_{A} \times U}\) of the capsule sequences is multiplied with the activated cosine similarity matrix to obtain the feature map \(\mathbf{F}_{g}^{R} \in \mathbb{R}^{V \times O_{A} \times U}\) \emph{i.e.},
\begin{equation}
\label{equ:retrieval}
\begin{array}{l}
\mathbf{F}_{g}^{R} = \mathrm{Split}(\mathbf{S}_{g}) \times \mathbf{\widehat{E}}_{g},
\end{array}
\end{equation}where $\mathrm{Split}(\cdot)$ represents the operation of split the final dimension along the channel axis.

The obtained relational sequences \(\mathbf{F}_{g}^{R}\) is integrated with learned mamba sequences $\mathbf{F}_{g}^{M}$ to obtain the sequence $\mathbf{F}_{g}^{C}$. Finally, four sequence $\mathbf{F}_{g}^{C}$ are integrated under multiple scanning directions through flip and transpose operations to ensure consistency with the first scanning direction
\begin{equation}
\label{equ:fusefour}
\begin{array}{l}
\mathbf{F}^{D} = \mathrm{MSA}(\mathrm{Cat}(\mathbf{F}_{1}^{C},\Psi\left(\mathbf{F}_{2}^{C}\right),\Gamma\left(\mathbf{F}_{3}^{C}\right),\Psi(\Gamma\left(\mathbf{F}_{4}^{C}\right)))),
\end{array}
\end{equation}where $\mathrm{MSA}(\cdot)$ means multi-head self-attention. $\Psi(\cdot)$ and $\Gamma(\cdot)$ represent the operations of transpose and flipping, respectively.

\subsection{Transformer Decoder}

In the decoder, following the idea of VST \cite{liu20214702}, two designed task-related tokens (\emph{i.e.}, a camouflage token \(\mathbf{t}^{c} \in \mathbb{R}^{1 \times d}\) and a boundary token \(\mathbf{t}^{b} \in \mathbb{R}^{1 \times d}\)) are added on the obtained tokens \(\mathbf{F}^{D}\) for completing camouflaged object segmentation and edge detection distinctively. Then, the all tokens are processed via transformer layers to capture global dependencies. In every layer, a patch-task-attention between \(\mathbf{F}_{j}^{D}\) and \(\mathbf{t}_{j}^{c}\) is designed for camouflage prediction $\mathbf{F}_{j}^{c}$, where \(j \in \lbrack 0,1,2\rbrack\) indicates the index of blocks in the decoder. $\mathbf{F}_{j}^{D}$ is mapped to queries $\mathbf{Q}_{j}^{c} \in \mathbb{R}^{l_{j} \times d_{j}}$ and $\mathbf{t}_{j}^{c}$ is mapped to a key $\mathbf{k}_{j}^{c} \in \mathbb{R}^{1 \times d_{j}}$ and a value $\mathbf{v}_{i}^{c} \in \mathbb{R}^{1 \times d_{j}}$, where $l_{j}$ and $d_{j}$ mean the length of the sequence and the channel dimension of the token, respectively. The camouflage prediction $\mathbf{F}_{j}^{c}$ can be computed by 
\begin{equation}
\label{equ:camouflageprediction}
\begin{array}{l}
\mathbf{F}_{j}^{c} = \mathrm{Sigmoid}\left( \mathbf{Q}_{j}^{c}{{\times \mathbf{k}}_{j}^{c}}^{T}/\sqrt{d} \right) \times \mathbf{v}_{j}^{c} \oplus \mathbf{F}_{j}^{D},
\end{array}
\end{equation}where $(\cdot)^T$ and $\oplus$ represent the transpose operation of the matrix and the operation of element-wise addition, respectively. In a similar way, for boundary prediction, \(\mathbf{F}_{j}^{D}\) is mapped to queries \(\mathbf{Q}_{j}^{b}\ \)and \(\mathbf{t}_{j}^{b}\) is mapped to a key \(\mathbf{k}_{j}^{b}\) and a value \(\mathbf{v}_{j}^{b}\) to gain the result
\begin{equation}
\label{equ:boundaryprediction}
\begin{array}{l}
\mathbf{F}_{j}^{b} = \mathrm{Sigmoid}\left( \mathbf{Q}_{j}^{b}{{\times \mathbf{k}}_{j}^{b}}^{T}/\sqrt{d} \right) \times \mathbf{v}_{j}^{b} \oplus \mathbf{F}_{j}^{D}.
\end{array}
\end{equation}

Whereafter, two linear transformations are applied with the sigmoid activation to project \(\mathbf{F}_{j}^{c}\), \(\mathbf{F}_{j}^{b}\) to scalars in {[}0, 1{]}. Therefore, get the final 2D camouflaged map \(\mathbf{\widetilde {F}}_{j}^{c}\) and boundary map  \(\mathbf{\widetilde{F}}_{j}^{b}\) at the corresponding scale.

\subsection{Loss Function}

In this work, both the weighted binary cross-entropy (BCE) loss function (\(l_{wbce}\)) and the Intersection over Union (IoU) loss (\(l_{iou}\)) \cite{Yu2016iou} are adopted as loss functions to train the network. Suppose $\mathbf{\widetilde {F}}^{c}$, $\mathbf{\widetilde {F}}^{b}$, $\mathbf{G}^{c}$ and $\mathbf{G}^{b}$ are the predicted camouflaged map, boundary map, corresponding camouflaged 
and boundary ground truth, respectively. $l_{wbce}$ can be expressed in the formula as follows:
\begin{equation}
\label{equ:wbceloss}
\begin{array}{l}
l_{wbce} = \sum_{j}{\left[l_{bce}\left( \mathbf{\widetilde {F}}_{j}^{c},\mathbf{G}_{j}^{c} \right) + l_{bce}\left(\mathbf{\widetilde {F}}_{j}^{b},\mathbf{G}_{j}^{b}\right)\right] \times w_{j}},
\end{array}
\end{equation}where $j$ is the index of blocks in the decoder and \(w_{j}\) is a set of hyperparameters that we set the value of \(\lbrack w_{0},w_{1},w_{2},w_{3}\rbrack\) to \(\lbrack 1,\ 0.8,\ 0.5,\ 0.5\rbrack\). In Eq. (\ref{equ:wbceloss})
\begin{equation}
\label{equ:bceloss}
\begin{array}{l}
l_{bce} = - \frac{1}{n} \sum_{m}\mathbf{G}_{m}log(\mathbf{\widetilde{F}}_{m}) + (1 - \mathbf{G}_{m})log(1 - \mathbf{\widetilde{F}}_{m}),
\end{array}
\end{equation}where \(m\) represents the pixel index and $n$ means the total number of pixels.

$l_{iou}$ is defined on the input scale,
\begin{equation}
\label{equ:iouloss}
\begin{array}{l}
l_{iou} = 1 - \frac{\sum_{m}{\mathbf{\widetilde {F}}{c}(m)\mathbf{G}^{c}(m)}}{\sum_{m}^{}{\lbrack \mathbf{\widetilde {F}}{c}(m) + \mathbf{G}^{c}(m)\  - \ \mathbf{\widetilde {F}}{c}(m)\mathbf{G}^{c}(m)\rbrack}}.
\end{array}
\end{equation}

\section{Experiment and Analysis}
\label{sec:Experiment}

In this section, we will carry out abundant experiments and analysis to provide a comprehensive understanding of the proposed method.

\subsection{Experimental Settings}

\textbf{Dataset.} We evaluate the proposed method on three widely public benchmarks.

CAMO \cite{le201945} is the first COD dataset, containing 1,250 camouflaged images with 1,000 training images and 250 testing images.

COD10K \cite{fan20202774} is a currently large COD datasets, consisting of 3,040 training images and 2,026 testing images..

NC4K \cite{lv202111586} is a recently released large-scale COD dataset containing 4,121 images. 

To ensure consistency with previous studies \cite{fan20226024}, 3040 samples from COD10K and 1000 samples from CAMO  are utilized as the training set, while the test set consisting of 2026 test images in COD10K, 250 test images in CAMO and the entire NC4K dataset.

\begin{table*}[!htbp]
    \centering	
    \caption{Quantitative comparison with 25 SOTA methods on three benchmark datasets. Notes ↑ / ↓ denote the larger/smaller is better, respectively. “–” is not available. The best and second best are \textbf{bolded} and \underline{underlined} for highlighting, respectively.}
    \begin{adjustbox}{max width=\textwidth}
    \setlength{\tabcolsep}{4pt}
    \footnotesize
			\begin{tabular}{c|cccc|cccc|cccc}
				\hline
				\multirow{2}{*}{\textbf{Method}}&
				\multicolumn{4}{c|}{\textbf{CAMO (250 images)}}&
				\multicolumn{4}{c|}{\textbf{COD10K (2026 images)}}&\multicolumn{4}{c}{\textbf{NC4K (4121 images)}}\cr\cline{2-13}
				&$MAE \downarrow$&$F_m \uparrow$&$E_m \uparrow$&$S_m \uparrow$&$MAE \downarrow$&$F_m \uparrow$&$E_m \uparrow$&$S_m \uparrow$&$MAE \downarrow$&$F_m \uparrow$&$E_m \uparrow$&$S_m \uparrow$\cr
				\hline
                      \multicolumn{13}{c}{\textbf{CapsNet based method}}\cr
                \hline
\bf{POCINet \cite{liu20215154}}&0.110&0.662&0.777&0.7017&0.051&0.0614&0.825&0.751&——&——&——&——\cr
                    \hline
                    \multicolumn{13}{c}{\textbf{CNNs based methods}}\cr
                    \hline
\bf{SINet \cite{fan20202774}}&0.091&0.708&0.829&0.746&0.042&0.691&0.874&0.777&0.058&0.775&0.883&0.810\cr
\bf{MGL \cite{zhai202112997mutual}}&0.089&0.725&0.811&0.772&0.035&0.709&0.852&0.815&0.053&0.782&0.868&0.832\cr
\bf{PFNet \cite{mei20218772}}&0.085&0.758&0.855&0.782&0.039&0.725&0.891&0.800&0.053&0.799&0.899&0.829\cr
\bf{LSR \cite{lv202111586}}&0.080&0.753&0.854&0.787&0.037&0.732&0.892&0.805&0.048&0.815&0.907&0.839\cr
\bf{C2FNet \cite{sun20211025}}&0.079&0.770&0.864&0.796&0.036&0.743&0.900&0.813&0.049&0.810&0.904&0.840\cr
\bf{UJSC \cite{li202110066}}&0.073&0.779&0.873&0.800&0.035&0.738&0.891&0.809&0.046&0.816&0.906&0.841\cr
\bf{SLTNet \cite{Cheng202213864SLT}}&0.082&0.763&0.848&0.792&0.036&0.681&0.875&0.804&0.049&0.787&0.886&0.830\cr
\bf{OCENet \cite{liu20222613}}&0.080&0.777&0.865&0.802&0.033&0.764&0.906&0.827&0.045&0.832&0.913&0.853\cr
\bf{BSANet \cite{zhu20223608}}&0.079&0.763&0.851&0.794&0.034&0.738&0.890&0.818&0.048&0.817&0.897&0.841\cr
\bf{FAPNet \cite{tao20227036}}&0.076&0.792&0.880&0.815&0.036&0.758&0.902&0.822&0.047&0.826&0.911&0.851\cr
\bf{BGNet \cite{sun20221335}}&0.073&0.799&0.882&0.812&0.033&0.774&0.916&0.831&0.044&0.833&0.916&0.851\cr
\bf{SegMaR \cite{jia20224703}}&0.071&0.803&0.884&0.816&0.034&0.775&0.907&0.833&0.046&0.827&0.907&0.841\cr
\bf{SINet-v2 \cite{fan20226024}}&0.070&0.801&0.895&0.820&0.037&0.752&0.906&0.815&0.047&0.823&0.914&0.847\cr
\bf{FDCOD \cite{zhong20224494}}&0.062&0.809&0.898&\underline{0.844}&0.030&0.749&0.918&0.837&0.052	&0.784&0.894&0.834\cr
\bf{ZoomNet \cite{pang20222160}}&0.066&0.805&0.892&0.820&0.029&0.780&0.911	&0.839&0.043&0.828&0.912&0.853\cr
\bf{R-MGL-v2 \cite{zhai20231897}}&0.086&0.731&0.847&0.769&0.034&0.733&0.879&0.816&0.050&0.801&0.899&0.838\cr
\bf{PopNet \cite{wu20231032}}&0.077&0.784&0.859&0.808&\underline{0.028}&0.786&0.910&\underline{0.851}&0.042&0.833&0.909&0.861\cr
\bf{FEDER \cite{he202322046}}&0.071&0.789&0.873&0.802&0.032&0.768&0.905&0.822&0.044	&0.833&0.915&0.847\cr
\bf{DGNet \cite{ji202391}}&\underline{0.057}&\underline{0.822}&\textbf{0.915}&0.839&0.033&0.759&0.911&0.823&0.042
&0.833&0.922&0.857\cr
\bf{DINet \cite{zhou7114decoupling}}&0.068&0.807&0.886&0.821&0.031&0.780&0.914&0.832&0.043
&0.839&0.919&0.856\cr
                    \hline
                    \multicolumn{13}{c}{\textbf{Transformer based methods}}\cr
                    \hline
\bf{UGTR \cite{yang20214126}}&0.086&0.754&0.855&0.785&0.035&0.742&0.891&0.818&0.051&0.807&0.899&0.839\cr
\bf{OSFormer \cite{pei20221937}}&0.073&0.767&0.858&0.799&0.034&0.701&0.881&0.811&0.049&0.790&0.891&0.832\cr
\bf{FDCOD \cite{zhong20224494}}&0.062&0.809&0.898&0.844&0.030&0.749&0.918&0.837&0.052	&0.784&0.894&0.834\cr
\bf{VSCode \cite{luo202417169}}&0.060&0.818&0.908&0.836&0.029&\underline{0.795}&\textbf{0.925}&0.847&\underline{0.038}&\underline{0.853}&\textbf{0.930}&\underline{0.874}\cr
\hline
\bf{MCRNet (Ours)}&\textbf{0.054}&\textbf{0.847}&\textbf{0.915}&\textbf{0.854}&\textbf{0.026}&\textbf{0.807}&\underline{0.924}&\textbf{0.854}&\textbf{0.036}&\textbf{0.857}&\textbf{0.930}&\textbf{0.875}\cr
\hline
			\end{tabular}

    \end{adjustbox}
    \vspace{-3mm}
\label{tab:sota}
\end{table*}

\textbf{Evaluation metrics.} Four commonly-used metrics are employed for COD task to assess model performance, including Mean Absolute Error $(MAE)$ \cite{achanta20091597}, maximum F-measure \(F_{m}\) \cite{margolin2014248}, maximum
enhanced-alignment measure \(E_{m}\) \cite{fan2018698} and structure-measure \(S_{m}\) \cite{fan20174558}.
Given a continuous camouflaged map, a binary mask $\hat F$ is achieved by thresholding the camouflaged map $F$. Precision is defined as $Precision = {{\left| {\hat F \cap G} \right|} \mathord{\left/
 {\vphantom {{\left| {\hat F \cap G} \right|} {\left| \hat F \right|}}} \right.
 \kern-\nulldelimiterspace} {\left| \hat F \right|}}$, and recall is defined as $Recall = {{\left| {\hat F \cap G} \right|} \mathord{\left/
 {\vphantom {{\left| {\hat F \cap G} \right|} {\left| G \right|}}} \right.
 \kern-\nulldelimiterspace} {\left| G \right|}}$.

$MAE$ is defined as
\begin{equation}
\label{equ:MAE}
MAE = \frac{1}{{\hat W \times \hat H}}{\sum\limits_{i} {\left| {F\left( {i} \right) - G\left( {i} \right)} \right|} } ,
\end{equation}
where $\hat W$ and $\hat H$ are the width and height of the image, respectively.

Maximum F-measure ($F_{m}$) is the maximum value of the F-measure $(F_\beta)$ under different thresholds. F-measure $(F_\beta)$ is an overall performance indicator, which is computed by
\begin{equation}
\label{equ:F-measure}
{F_\beta } = \frac{{\left( {1 + {\beta ^2}} \right)Precision \times Recall}}{{{\beta ^2}Precision + Recall}}.
\end{equation}
As suggested in \cite{margolin2014248}, ${{\beta ^2} = 0.3}$. 

Maximum enhanced-alignment measure ($E_m$) is the maximum value of E-measure under different thresholds, which combines local pixel values with the image-level mean value to jointly evaluate the similarity between the prediction and the ground truth.

Structure-measure ($S_m$) is computed by
\begin{equation}
\label{equ:S}
{S_m} = \alpha {S_o} + \left( {1 - \alpha } \right){S_r},
\end{equation}
where $S_o$ and $S_r$ represent the object-aware and region-aware structure similarities between the prediction and the ground truth, respectively. $\alpha$ is set to 0.5 \cite{fan20174558}.

\textbf{Implementation details.}
The proposed MCRNet is implemented by PyTorch. The tiny Swin-transformer \cite{liu20219992} is adopted as the network encoder. Other modules are randomly initialized. Each image is resized to 384×384 pixels and then randomly cropped to 352×352 for training. The network employs the Adam optimizer \cite{diederik2014} with an initial learning rate of 0.0001, which is reduced by a factor of 10 at half and three-quarters of the total training steps. The complete training process contains a total of 150,000 training steps with a batch size of 8 using a 4090 GPU.

\begin{figure*}[htbp]
\centering
 \includegraphics[width=1\linewidth]{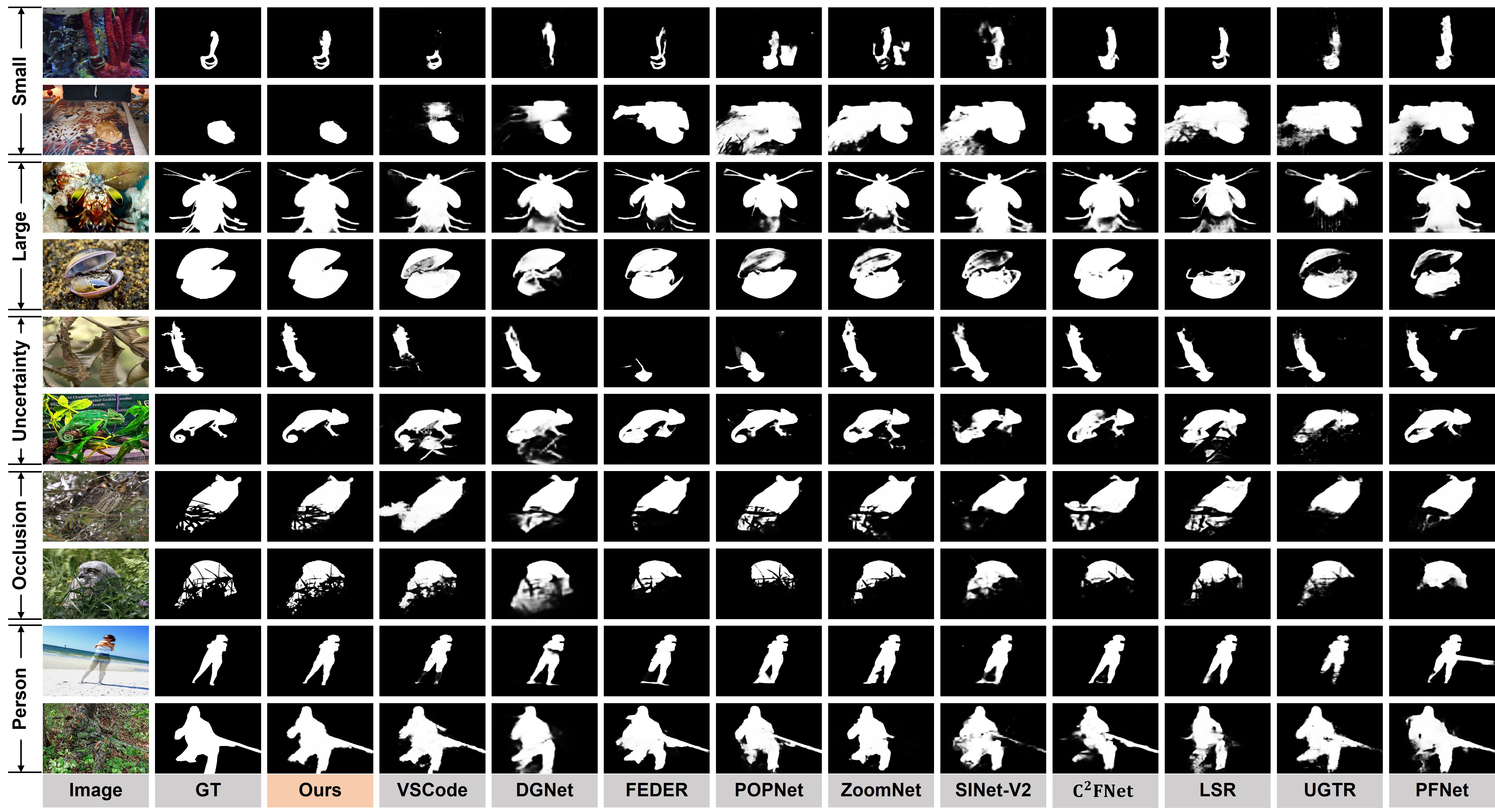}
\caption{Visual comparisons of the proposed MCRNet and other popular SOTA methods. The  proposed MCRNet segments the camouflaged objects well in challenging scenes, including small objects, large objects, the objects with uncertain boundaries, the objects that are obscured, and the concealed persons.}
\label{fig:vision_result}
\end{figure*}

\begin{table*}[!htbp]
    \centering	
    \caption{Ablation study. ``B" denotes the baseline of the Swin Transformer-T backbone. ``MCG"
 indicates the incorporation of mamba capsules routing into the baseline. ``CSDR" represents the capsules spatial details retrieval.}
    \setlength{\tabcolsep}{2pt}
    \begin{adjustbox}{max width=\textwidth}
    \footnotesize
			\begin{tabular}{cccc|cccc|cccc|cccc}
				\hline
				\multicolumn{4}{c|}{\textbf{Candidate}}&
				\multicolumn{4}{c|}{\textbf{CAMO (250 images)}}&
				\multicolumn{4}{c|}{\textbf{COD10K (2026 images)}}&\multicolumn{4}{c}{\textbf{NC4K (4121 images)}}\cr
                \hline
				&B&MCG&CSDR&$MAE \downarrow$&$F_m \uparrow$&$E_m \uparrow$&$S_m \uparrow$&$MAE \downarrow$&$F_m \uparrow$&$E_m \uparrow$&$S_m \uparrow$&$MAE \downarrow$&$F_m \uparrow$&$E_m \uparrow$&$S_m \uparrow$\cr
				\hline
            
(a)&\checkmark&&&0.0615&0.8250&0.9087&0.8398&0.0292&0.7921&0.9181&0.8449&0.0404&0.8435&0.9242&0.8663\cr
(b)&\checkmark&\checkmark&&0.0581&0.8338&0.9107&0.8412&0.0269&0.8053&0.9221&0.8507&0.0382&0.8540&0.9287&0.8710\cr
(c)&\checkmark&\checkmark&\checkmark&\textbf{0.0547}&\textbf{0.8466}&\textbf{0.9151}&\textbf{0.8536}&\textbf{0.0264}&\textbf{0.8069}&\textbf{0.9236}&\textbf{0.8544}&\textbf{0.0366}&\textbf{0.8570}&\textbf{0.9301}&\textbf{0.8751}\cr
\hline
			\end{tabular}
    \end{adjustbox}
\label{tab:abla_module}
\end{table*}

\subsection{Comparison with the State-of-the-arts}

In order to demonstrate the efficacy of the proposed method, the comparative analysis is conducted with 25 recent state-of-the-art methodologies, including the capsule network based method (\emph{i.e.}, POCINet \cite{liu20215154}), the CNNs based methods ( SINet \cite{fan20202774}, MGL \cite{zhai202112997mutual}, PFNet \cite{mei20218772}, LSR \cite{lv202111586}, C2FNet \cite{sun20211025}, UJSC \cite{li202110066}, R-MGL-v2 \cite{zhai20231897}, SLTNet \cite{Cheng202213864SLT}, OCENet \cite{liu20222613}, BSANet \cite{zhu20223608}, FAPNet \cite{tao20227036}, BGNet \cite{sun20221335}, SegMaR \cite{jia20224703}, SINet-v2 \cite{fan20226024}, FDCOD \cite{zhong20224494}, ZoomNet \cite{pang20222160}, PopNet \cite{wu20231032}, FEDER \cite{he202322046}, DGNet \cite{ji202391}, DINet\cite{zhou7114decoupling}) and the transformer based methods (UGTR \cite{yang20214126}, OSFormer \cite{pei20221937}, FDCOD \cite{zhong20224494}, VSCode \cite{luo202417169}).  For a fair comparison, all the predictions of these methods are either provided by the authors or generated by models retrained based on the open source codes with the same code. 

\textbf{Quantitative analysis.} Table \ref{tab:sota} presents a summary of the quantitative analysis of the proposed approach in contrast to 25 rivals on three COD datasets in terms of four evaluation metrics. From the metric data, it can be seen that the proposed MCRNet comprehensively surpasses all existing state-of-the-art methods. Compared to the second best and transformer based network called VSCode \cite{luo202417169}, the method achieves average performance gains of 8.5\%, 1.7\%, 0.2\%, 1.0\% in terms of $MAE$, \(F_{m}\), \(E_{m}\), \(S_{m}\), respectively after averaging all metrics of these three datasets. Compared with FEDER \cite{he202322046} based on CNN, which also completes object segmentation and edge detection tasks, it shows significant performance improvements of 21.1\%, 5.1\%, 2.9\%, and 4.5\% respectively in the four indicators  from the average perspective. Compared to the multi-scale method ZoomNet \cite{pang20222160}, the average gains are 15.9\%, 4.1\%, 2.0\%, and 2.8\%, respectively. Besides, compared with the CapsNets based method POCINet \cite{liu20215154}, we have achieved a significant improvement in segmentation accuracy, which benefits from the mamba capsule routing in our MCRNet.

\begin{table*}[!htbp]
    \centering	
    \caption{Ablation study on different type-level capsules generations. ``FC" indicates the fully connection operation, and ``Mamba" represents our method.}
    \setlength{\tabcolsep}{4pt}
    \begin{adjustbox}{max width=\textwidth}
    \footnotesize
			\begin{tabular}{c|cccc|cccc|cccc}
				\hline
				\multirow{2}{*}{\textbf{Generation}}&
				\multicolumn{4}{c|}{\textbf{CAMO (250 images)}}&
				\multicolumn{4}{c|}{\textbf{COD10K (2026 images)}}&\multicolumn{4}{c}{\textbf{NC4K (4121 images)}}\cr
                \cline{2-13}
				&$MAE \downarrow$&$F_m \uparrow$&$E_m \uparrow$&$S_m \uparrow$&$MAE \downarrow$&$F_m \uparrow$&$E_m \uparrow$&$S_m \uparrow$&$MAE \downarrow$&$F_m \uparrow$&$E_m \uparrow$&$S_m \uparrow$\cr
				\hline
            
\textbf{FC}&0.0581&0.8338&0.9107&0.8412&0.0269&0.8050&0.9202&0.8500&0.0382&0.8540&0.9216&0.8705\cr
\hline
\textbf{Mamba}&\textbf{0.0547}&\textbf{0.8466}&\textbf{0.9151}&\textbf{0.8536}&\textbf{0.0264}&\textbf{0.8069}&\textbf{0.9236}&\textbf{0.8544}&\textbf{0.0366}&\textbf{0.8570}&\textbf{0.9301}&\textbf{0.8751}\cr
\hline
			\end{tabular}
    \end{adjustbox}
\label{tab:abla_generation}
\end{table*}

\begin{table*}[!htbp]
    \centering	
    \caption{Ablation study on the number of mamba capsules. All models are trained based on Table \ref{tab:abla_module} (c).}
    \setlength{\tabcolsep}{4pt}
    \begin{adjustbox}{max width=\textwidth}
    \footnotesize
			\begin{tabular}{c|cccc|cccc|cccc}
				\hline
				\multirow{2}{*}{\textbf{Number}}&
				\multicolumn{4}{c|}{\textbf{CAMO (250 images)}}&
				\multicolumn{4}{c|}{\textbf{COD10K (2026 images)}}&\multicolumn{4}{c}{\textbf{NC4K (4121 images)}}\cr
                \cline{2-13}
				&$MAE \downarrow$&$F_m \uparrow$&$E_m \uparrow$&$S_m \uparrow$&$MAE \downarrow$&$F_m \uparrow$&$E_m \uparrow$&$S_m \uparrow$&$MAE \downarrow$&$F_m \uparrow$&$E_m \uparrow$&$S_m \uparrow$\cr
				\hline
            
\textbf{0}&0.0615&0.8250&0.9087&0.8398&0.0292&0.7921&0.9181&0.8449&0.0404&0.8435&0.9242&0.8663\cr
\textbf{32}&0.0547&\textbf{0.8466}&\textbf{0.9151}&\textbf{0.8536}&\textbf{0.0264}&\textbf{0.8069}&\textbf{0.9236}&\textbf{0.8544}&0.0366&0.8570&0.9301&0.8751\cr
\textbf{64}&0.0548&0.8421&0.9127&0.8489&0.0266&0.8036&0.9217&0.8522&\textbf{0.0365}&\textbf{0.8592}&\textbf{0.9315}&\textbf{0.8760}\cr
\textbf{96}&\textbf{0.0545}&0.8455&0.9136&0.8500&0.0268&0.8045&0.9229&0.8529&0.0371&0.8570&0.9309&0.8746\cr
\textbf{128}&0.0575&0.8372&0.9105&0.8470&0.0272&0.8035&0.9224&0.8511&0.0373&0.8557&0.9292&0.8731\cr
\hline
			\end{tabular}
    \end{adjustbox}
\label{tab:abla_number_capsule}
\end{table*}

\textbf{Qualitative analysis.} Fig. \ref{fig:vision_result} presents the segmentation visualizations of the MCRNet with ten good methods. From the three test sets, camouflage objects of diverse sizes and camouflage scenes of various types are selected for visualizations, encompassing small objects, large objects, the objects with uncertain boundaries, the objects that are obscured, and the concealed persons. As can be witnessed from the first and second rows, the small camouflage objects can be detected extremely well and not missed, particularly the second row of camouflage objects with a high resemblance to the background. In the instance of large camouflaged objects, they can be identified with good completeness. In the case of fuzzy boundaries, the camouflaged object can be detected entirely from the low-contrast background. It is worthy of mention that for objects obscured by the jungle, they can be distinguished meticulously. Similarly a favorable detection effect has also been attained in the scene  containing persons. The aforementioned outstanding segmentation of camouflaged objects are attributed to the exploration of the part-whole relationship by the proposed MCRNet.

\subsection{Ablation Analysis}

\begin{table*}[!htbp]
    \centering	
    \caption{Ablation study on different scanning directions for capsules sequence.}
    \setlength{\tabcolsep}{2.5pt}
    \begin{adjustbox}{max width=\textwidth}
    \footnotesize
			\begin{tabular}{c|cccc|cccc|cccc}
				\hline
				\multirow{2}{*}{\textbf{Capsule Sequence Order}}&
				\multicolumn{4}{c|}{\textbf{CAMO (250 images)}}&
				\multicolumn{4}{c|}{\textbf{COD10K (2026 images)}}&\multicolumn{4}{c}{\textbf{NC4K (4121 images)}}\cr
                \cline{2-13}
				&$MAE \downarrow$&$F_m \uparrow$&$E_m \uparrow$&$S_m \uparrow$&$MAE \downarrow$&$F_m \uparrow$&$E_m \uparrow$&$S_m \uparrow$&$MAE \downarrow$&$F_m \uparrow$&$E_m \uparrow$&$S_m \uparrow$\cr
				\hline
            
\textbf{One Direction}&0.0572&0.8396&0.9111&0.8463&0.0268&0.8040&0.9236&0.8527&0.0370&0.8567&0.9301&0.8748\cr
\textbf{Two Directions}&0.0561&0.8389&0.9097&0.8470&\textbf{0.0262}&0.8063&\textbf{0.9239}&0.8542&0.0373&0.8555&\textbf{0.9301}&0.8741\cr
\hline
\textbf{Four Directions (Ours)}&\textbf{0.0547}&\textbf{0.8466}&\textbf{0.9151}&\textbf{0.8536}&0.0264&\textbf{0.8069}&0.9236&\textbf{0.8544}&\textbf{0.0366}&\textbf{0.8570}&\textbf{0.9301}&\textbf{0.8751}\cr
\hline
			\end{tabular}
    \end{adjustbox}
\label{tab:abla_number_direction}
\end{table*}

\begin{table*}[!htbp]
    \centering	
    \caption{FLOPs, Parameters and Time of different capsule routing algorithms for part-whole relational COD.}
    \setlength{\tabcolsep}{0.25pt}
    \begin{adjustbox}{max width=\textwidth}
    \footnotesize
			\begin{tabular}{c|c|c|c|cccc|cccc|cccc}
				\hline
				\multirow{2}{*}{\textbf{Network}}&
				\multicolumn{1}{c|}{\textbf{FLOPs}}&
				\multicolumn{1}{c|}{\textbf{Params}}&
                \multicolumn{1}{c|}{\textbf{Time}}&
				\multicolumn{4}{c|}{\textbf{CAMO (250 images)}}&
				\multicolumn{4}{c|}{\textbf{COD10K (2026 images)}}&\multicolumn{4}{c}{\textbf{NC4K (4121 images)}}\cr
                \cline{5-16}
				&(G) $\downarrow$&(M) $\downarrow$&(s) $\downarrow$ &$MAE \downarrow$&$F_m \uparrow$&$E_m \uparrow$&$S_m \uparrow$&$MAE \downarrow$&$F_m \uparrow$&$E_m \uparrow$&$S_m \uparrow$&$MAE \downarrow$&$F_m \uparrow$&$E_m \uparrow$&$S_m \uparrow$\cr
				\hline
            
\textbf{\makecell{EM Routing\\ \cite{hinton2018matrix}}}&155.16&77.96&0.039&0.0623&0.8269&0.9045&0.8378&0.0292&0.7931&0.9168&0.8453&0.0395&0.8481&0.9250&0.8686\cr
\hline
\textbf{\makecell{DCR\\ \cite{liu20226719}}}&150.37&73.62&0.036&0.0599&0.8310&0.9101&0.8425&0.0279&0.7967&0.9210&0.8477&0.0384&0.8511&0.9284&0.8699\cr
\hline
\textbf{\makecell{MCR\\ (Ours)}}&\textbf{145.74}&\textbf{69.11}&\textbf{0.028}&\textbf{0.0547}&\textbf{0.8466}&\textbf{0.9151}&\textbf{0.8536}&\textbf{0.0264}&\textbf{0.8069}&\textbf{0.9236}&\textbf{0.8544}&\textbf{0.0366}&\textbf{0.8570}&\textbf{0.9301}&\textbf{0.8751}\cr
\hline

			\end{tabular}
    \end{adjustbox}
\label{tab:abla_different_capsule}
\end{table*}

\begin{figure}[t]
\centering
 \includegraphics[width=1\linewidth]{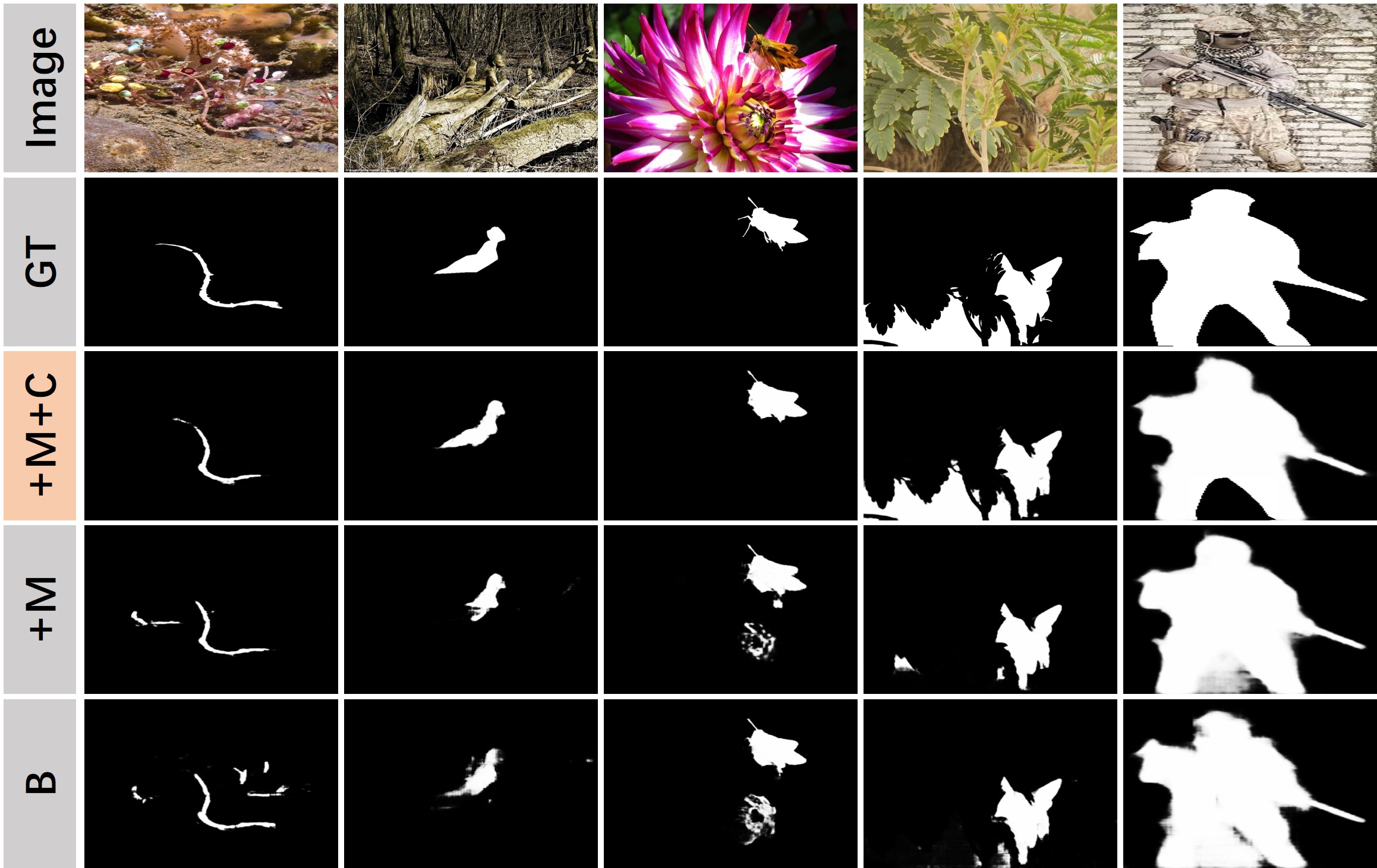}
\caption{Visual comparisons for the ablation of MCG and CSDR. ``B", ``M" and ``C" stand for Baseline, MCG and CSDR, respectively. ``+M+C" marked in orange represents the entire MCRNet.}
\label{fig:ablation_vision_result}
\end{figure}

\textbf{Effectiveness of MCG and CSDR.} The proposed MCG and CSDR module exert a significant role in facilitating proposed MCRNet for part-whole relational camouflaged object detection. To dig into the contributions of these two components, we design ablation studies by removing them from the entire framework. Table \ref{tab:abla_module} and Fig. \ref{fig:ablation_vision_result} demonstrate the performance and visualizations for the ablation study. Comparing the fourth and fifth rows in Fig. \ref{fig:ablation_vision_result}, it can be observed that MCG enables to better separate the camouflage object from its surroundings, which is also proven in Table \ref{tab:abla_module} (a) and (b) in terms of performance. This is attributed to the mamba capsules by the latent state mechanism in the selective SSM \cite{gu20242312} model that realizes the modeling of global spatial structure information. Comparing the third and fourth rows in Fig. \ref{fig:ablation_vision_result}, it can be seen that our CSDR can effectively detect what other models cannot and enhance the integrity of camouflaged object. Similar conclusion can be achieved by comparing Table \ref{tab:abla_module} (b) and (c). This proves that the spatial details retrieved by CSDR help the segmentation of the camouflaged objects.

\textbf{Generation of type-level capsules.} To prove the validity of MCG module for type-level mamba capsules generation, we compare it with a straightforward manner that uses a linear mapping to generate the type-level capsules. As shown in Table \ref{tab:abla_generation}, capsules generated with mamba outperform those generated with the fully-connected layer. This superiority can be attributed to that the latent state using the selective SSM \cite{gu20242312} in VMamba \cite{liu2024arxiv} enables accumulates selected token information for comprehensive global context. By contrast, full connection simply implements a weighted sum of all tokens in the sequence without retaining spatial structure context well.

\textbf{Number of mamba capsules.} Table \ref{tab:abla_number_capsule} demonstrates the impact of the quantity of mamba capsules in the proposed network on the model's detection capability. As is widely acknowledged in Table \ref{tab:abla_number_capsule}, insufficient capsules undermine the characterization ability of camouflaged objects, while excessive capsules result in overfitting and a decline in detection performance. After conducting a qualitative analysis the optimal equilibrium in terms of detection performance and generalization ability is to set to 32 for mamba capsules number, which is the setting in this paper to facilitate the efficient experiments.

\textbf{Scanning direction for capsules sequence.} Table \ref{tab:abla_number_direction} explores various serialization directions to study the scanning directions in the MCRNet, including one direction, two directions and four directions, which can be referred to Fig. \ref{fig:MCG}. Specifically, one direction and two directions possess the scannings of 'Z' and 'Z' \& 'N', respectively. In Table \ref{tab:abla_number_direction}, the combination of four scannings achieves the best performance, which indicates the capability of various scanning directions for global context extraction.

\textbf{Efficiency analysis.} To explore the efficiency of the proposed MCRNet for the pipeline of part-whole relational COD based on CapsNets, we replace the proposed mamba capsule routing with some previous capsule routing algorithms, including EM routing \cite{hinton2018matrix} and Disentangled Capsule Routing (DCR) \cite{liu20226719} in the entire MCRNet framework. In Table \ref{tab:abla_different_capsule}, the proposed mamba capsule routing achieves the lowest FLOPs, and parameters, and highest inference speed, while performing best on various datasets, which demonstrates the complexity efficiency and performance superiority.

\subsection{Failure Cases}

\begin{figure}[t]
\centering
 \includegraphics[width=1\linewidth]{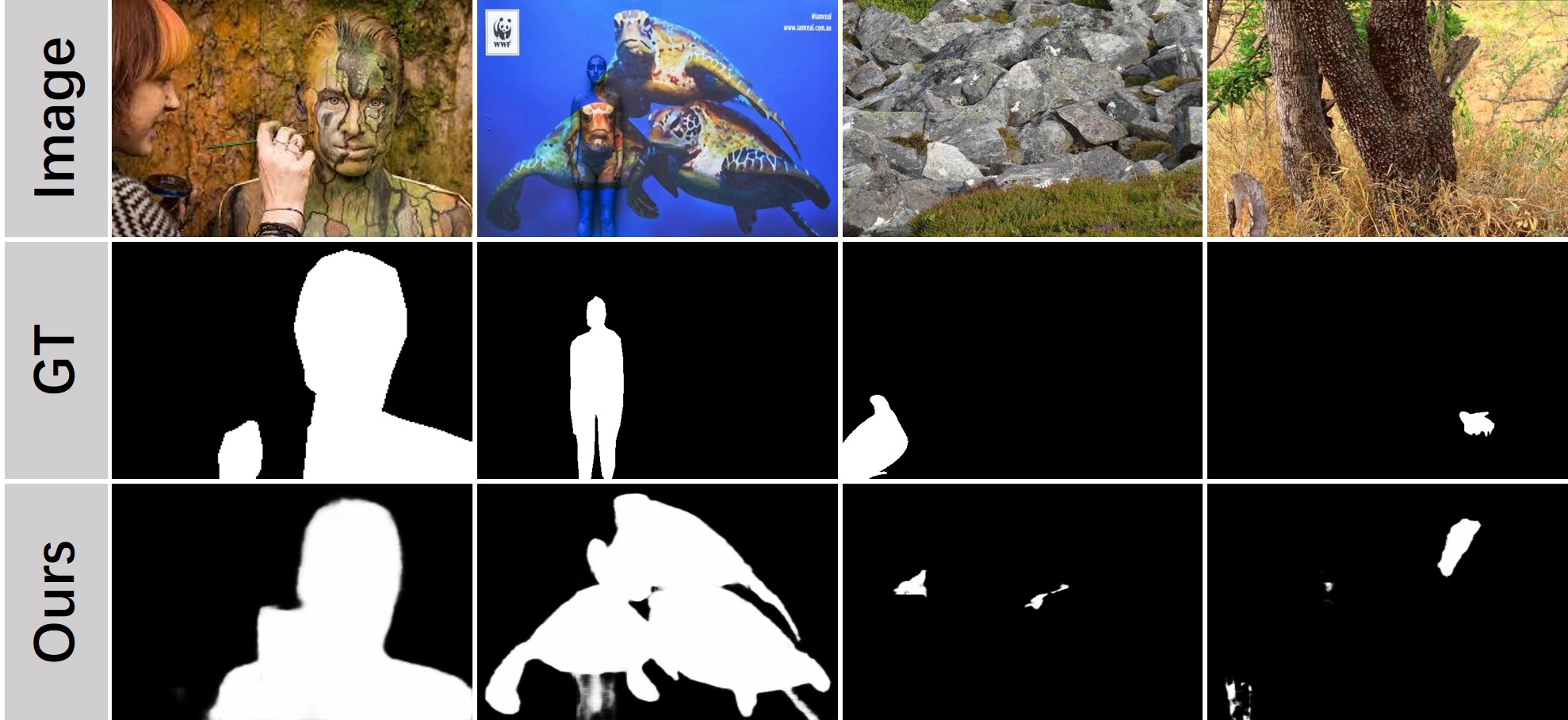}
\caption{Failure cases. From top to bottom: Images, GT, and
results of our method.}
\label{fig:fail}
\end{figure}

Fig. \ref{fig:fail} displays some failure cases of our camouflaged detector
on extremely complex scenes. For instance, as depicted in the left two columns of Fig. \ref{fig:fail}, the model's detection of camouflaged targets is influenced by salient target detection, leading to unnecessary object parts being detected and a shift in detection focus. Besides, in the scene shown in the right two columns of Fig. \ref{fig:fail}, subpar performance exhibits in detecting small objects out of the center due to increased observation angle distance, making it challenging to discern camouflage on small objects. In future endeavors, on the basis of leveraging part-whole relational methods, we will leverage the power of Large Language Model (LLM) \cite{Alec2018GPT} to help the understanding of the camouflaged scene for better concealed object searching and identification.

\section{Conclusions}
\label{sec:Conclusion}

In this paper, we have designed the Mamba Capsule Routing Network (MCRNet) for the pipeline of part-whole relational COD task. To achieve the lightweight of capsule routing for part-whole relationships exploration, a MCG was designed to generate the type-level mamba capsules from the pixel-level capsules, which ensures a lightweight capsule routing at the type level. On top of that, the CSDR module was designed to retrieve spatial details of the mamba capsules for the final camouflaged object detection. Extensive experiments have demonstrated that our proposed network module significantly enhances the detection performance of camouflaging objects. In future work, we will apply the LLM  \cite{Alec2018GPT} to improve the caouflaged object detection from the multi-modal understanding perspective.

\noindent\textbf{Data Availability Statement} 

The datasets used and analyzed during the current study are available in the following public domain resources:

\begin{itemize} 

\item \textbf{CAMO:} \href{https://github.com/ondyari/FaceForensics}{\textcolor{blue}{https://github.com/ondyari/FaceForensics}}

\item \textbf{COD10K:} \href{https://github.com/DengPingFan/SINet}{\textcolor{blue}{https://github.com/DengPingFan/SINet}}

\item \textbf{NC4K:} \href{https://github.com/JingZhang617/COD-Rank-Localize-and-Segment}{\textcolor{blue}{https://github.com/JingZhang617/COD-Rank-Localize-and-Segment}}

\end{itemize}

The models and source data generated and analyzed during the current study are available from the corresponding author upon reasonable request.


%
%


\bibliographystyle{spbasic}  
\bibliography{sn-bibliography}    

%
%

\end{sloppypar}
\end{document}